\documentclass{article}

\PassOptionsToPackage{numbers, compress}{natbib}

\usepackage[final]{neurips_2023}

\usepackage[utf8]{inputenc} % allow utf-8 input
\usepackage[T1]{fontenc}    % use 8-bit T1 fonts
\usepackage{hyperref}       % hyperlinks
\usepackage{url}            % simple URL typesetting
\usepackage{booktabs}       % professional-quality tables
\usepackage{amsfonts}       % blackboard math symbols
\usepackage{nicefrac}       % compact symbols for 1/2, etc.
\usepackage{microtype}      % microtypography
\usepackage{xcolor}         % colors

\usepackage[pdftex]{graphicx}
\usepackage[export]{adjustbox}
\usepackage{amsmath}

\title{Information theoretic study of the neural geometry induced by category learning}

\author{%
  Laurent Bonnasse-Gahot$^{1}$ \quad Jean-Pierre Nadal$^{1,2}$\\
  $^{1}$Centre d'Analyse et de Math\'ematique Sociales (CAMS, CNRS - EHESS)\\
  \'Ecole des Hautes \'Etudes en Sciences Sociales, Paris, France \\
  $^{2}$Laboratoire de Physique de l’\'Ecole normale supérieure, \\
  ENS, Universit\'e PSL, CNRS, Sorbonne Universit\'e, Universit\'e Paris Cit\'e, Paris, France\\
  \texttt{lbg@ehess.fr}, 
  \texttt{jean-pierre.nadal@phys.ens.fr} \\
}

\begin{document}

\maketitle
\vspace{-0.42cm}

\begin{abstract}
Categorization is an important topic both for biological and artificial neural networks. Here, we take an information theoretic approach to assess the efficiency of the representations induced by category learning. We show that one can decompose the relevant Bayesian cost into two components, one for the coding part and one for the decoding part. Minimizing the coding cost implies maximizing the mutual information between the set of categories and the neural activities. We analytically show that this mutual information can be written as the sum of two terms that can be interpreted as (i) finding an appropriate representation space, and, (ii) building a representation with the appropriate metrics, based on the neural Fisher information on this space. One main consequence is that category learning induces an expansion of neural space near decision boundaries. Finally, we provide numerical illustrations that show how Fisher information of the coding neural population aligns with the boundaries between categories.
\end{abstract}

\vspace{-0.32cm}
\section{Introduction}
\label{sec:intro}
The study of categorization is an important field of research both for biological and artificial neural networks. Here, we take an information theoretic approach to study the optimal properties of the neural representations induced by category learning. We extend the formalism introduced in~\cite{LBG_JPN_2008,LBG_JPN_2012} to the case of multilayer networks, allowing to cast the approach and results within the machine learning framework. We consider multilayer feedforward networks whose goal is to learn a categorization task. We first introduce the mean Bayes risk adapted to a categorization task. We show that the minimization of this cost amounts to dealing with two issues: optimizing the decision stage in order to provide the best possible estimator of the category given the neural activities; and optimizing the stimulus encoding (through the multilayer processing) by maximizing the mutual information between categories and neural code. 

We then characterize the mutual information between the discrete categories and the neural activities in a coding layer, in the limit of a high signal-to-noise ratio. This limit allows to reveal the neural metrics relevant for the categorization task. It shows that maximizing the mutual information leads to finding the feature space most relevant for the classification (and amenable to easy decoding), and to probe this space with a particular metric. The latter depends on a ratio between two Fisher information quantities: one that is specific to the neural coding, and one that quantifies the classification uncertainty. As a result of the optimization, the space will be expanded near a class boundary, and contracted far from a boundary. This implies a better ability to discriminate between nearby inputs in the vicinity of a class boundary, than far from such boundary. This effect, well studied in cognitive science, is called categorical perception~\citep{Harnad_1987}. Finally, we provide numerical experiments that illustrate how learning modifies the metrics defined by the neural activities coding for categories. In an Appendix we briefly discuss the links and differences with the information bottleneck approach~\citep{tishby99information,tishby2000information}.

\section{Revealing the metric of internal representations}
\label{sec:revealing}
\paragraph{Cost function: Decoupling into coding and decoding parts.}
We assume we are given a discrete set of classes/categories, $\mu=1,\ldots,M $. Each category is characterized by a density distribution $P(\mathbf{s}|\mu)$ over the input (sensory) space. A sensory input $\mathbf{s} \in \mathbb{R}^{N_0}$ elicits a cascade of neural responses (multilayer feedforward processing), $\mathbf{r}^{(1)} \in \mathbb{R}^{N_1}$, $\ldots$,  $\mathbf{r}^{(L)} \in \mathbb{R}^{N_L}$. Finally, an estimate $\widehat{\mu}$ of $\mu$  is extracted from the observation of the neural activity of the last coding layer $\mathbf{r}$, possibly through a decoding cascade of processing.\\
For a given stimulus $\mathbf{s}$ and a neural activity $\mathbf{r}$, the relevant Bayesian quality criterion is given by the divergence $\mathcal{C}(\mathbf{s},\mathbf{r})$ between the true probabilities $\{P(\mu|\mathbf{s}), \mu=1,...,M\}$ and the estimator $\{g_{\mu}(\mathbf{r}), \mu=1,...,M\}$, defined as the Kullback-Leibler divergence (or relative entropy) \citep{Cover_Thomas_2006}, $\mathcal{C}(\mathbf{s},\mathbf{r}) \equiv \sum_{\mu=1}^{M} P(\mu|\mathbf{s}) \ln \frac{P\mu|\mathbf{s})}{g_{\mu}(\mathbf{r})}$. Averaging over $\mathbf{r}$ given $\mathbf{s}$, and then over $\mathbf{s}$, one can show that the resulting mean Bayesian cost $\mathcal{\overline{C}}$ induced by the estimation can be written as:
\begin{equation}
	\mathcal{\overline{C}} 
	= \mathcal{\overline{C}}_{coding} + \mathcal{\overline{C}}_{decoding}
	\label{eq:cost_main_expression}
\end{equation}
with $\mathcal{\overline{C}}_{coding} \,=\, I[\mu,\mathbf{s}] - I[\mu,\mathbf{r}]$, where $I[X,Y]$ denotes the mutual information between the random variables $X$ and $Y$, and  $\mathcal{\overline{C}}_{decoding} \,=\, \int D_{KL}(P_{\mu|\mathbf{r}}||g_{\mu|\mathbf{r}})\;P(\mathbf{r}) \, d\mathbf{r}$, is the average of the Kullback-Leibler divergence of $P(\mu|\mathbf{r})$ from the network output $g_{\mu}(\mathbf{r})$. \\
The consequences of this decoupling are as follows.\\
\textit{(i) Optimal decoding.} The decoding cost $\mathcal{\overline{C}}_{decoding}$ is the average relative entropy between the true probability of the category given the neural activity, and the output $g$. It is the only term depending on $g$, hence the function minimizing the cost given by Eq.~(\ref{eq:cost_main_expression}) is (if it can be realized) $g_{\mu}(\mathbf{r}) = P(\mu|\mathbf{r})$. \\
\textit{(ii) Optimal coding.} The coding cost $\mathcal{\overline{C}}_{coding}$ is the difference between the information content of the signal and the mutual information between category membership and neural activity. Since processing cannot increase information \citep{Blahut_1987}, the information $I[\mu,\mathbf{r}]$ conveyed by $\mathbf{r}$ about $\mu$ is at most equal to the one conveyed by the sensory input $\mathbf{s}$: $I[\mu, \mathbf{s}] - I[\mu, \mathbf{r}] \; \geq 0$. If it is possible to find parameters such that the optimal estimator is reached, then the full average cost function (\ref{eq:cost_main_expression}) reduces to this difference in mutual information quantities. In such case, the cost function is minimized by maximizing the mutual information $I[\mu, \mathbf{r}]$ between neural activity and category membership. Hence the infomax principle is here an outcome of the global optimization problem. This result is somewhat related to the information bottleneck approach~\citep{tishby99information}, see Appendix for details.

\paragraph{Geometry of internal representations.}
\label{sec:geometry}
The whole chain of neural responses can be seen as first extracting a representation $\mathbf{x} \in \mathbb{R}^K$ over which $N$ neurons have their activities $\mathbf{r}$: the neurons form a population code covering a $K$-dimensional ($K \ll N$) feature space given by $\mathbf{x}$. This is motivated by the many recent works that show how (both biological and artificial) neural activities can be understood as acting on a lower-dimensional manifold (for works in neuroscience, see e.g. \citealp{archer2014low,sadtler2014neural,cunningham2014dimensionality,gallego2017neural,chung2021neural,jazayeri2021interpreting}, and for the machine learning literature, see e.g. \citealp{ma2018dimensionality, ansuini2019intrinsic, pope2021intrinsic}). Thus, one has the following Markov chain: $\mu \rightarrow \mathbf{s} \rightarrow \mathbf{x}=X(\mathbf{s}) \rightarrow \mathbf{r} \rightarrow \widehat{\mu}.$\\
For the decoding part, the minimization of the cost $\mathcal{\overline{C}}_{decoding}$ implies that, in this asymptotic limit of large signal-to-noise ratio (large number $N$ of coding cells),
$g_{\mu}(\mathbf{r}) = P(\mu|\mathbf{r})$ is an efficient estimator of $P(\mu|\mathbf{x})$: it is unbiased and saturates the associated Cramér-Rao bound \citep{LBG_JPN_2012}.\\
For the coding part, as we have seen, the minimization of the cost $\mathcal{\overline{C}}_{coding}$ leads to the maximization of the mutual information $I[\mu,\mathbf{r}]$ between the categories and the neural representation provided by the network prior to decoding. From the data processing theorem, we have that $I[\mu,\mathbf{r}] \leq  I[\mu, \mathbf{x}] \leq I[\mu,\mathbf{s}]$. Thus, for a given projection space $X$, at best $I[\mu,\mathbf{r}]=  I[\mu, \mathbf{x}]$, and optimization with respect to the choice of the space $X$ gives optimally $I[\mu, \mathbf{x}]=I[\mu,\mathbf{s}]$. The quality of the projection $X$ is given by how much the probability of the category given the stimulus is well approximated by the probability of the category given the projection $X(\mathbf{s})$. As for the mutual information between categories and neural code, in a regime of high signal-to-noise ratio, one can write \citep{LBG_JPN_2008}:
\begin{equation}
I[\mu,\mathbf{r}] = I[\mu,\mathbf{x}] - \frac{1}{2} \int  \operatorname{tr} \left( F_{\text{cat}}^{\,T}(\mathbf{x})\,F_{\text{code}}^{\,-1}(\mathbf{x}) \right)\; P(\mathbf{x}) \,d\mathbf{x} 
\label{eq:Delta_general}
\end{equation}
where $F_{\text{code}}(\mathbf{x})$ and $F_{\text{cat}}(\mathbf{x})$ are $K\times K$ Fisher information  matrices:\\
$\big[F_{\text{code}}(\mathbf{x})\big]_{ij} \; =  -  \int_r    \frac{\partial^2 \ln P(\mathbf{r}|\mathbf{x}) }{\partial x_i \partial x_j}\;P(\mathbf{r}|\mathbf{x}) \,d\mathbf{r}$, $\big[F_{\text{cat}}(\mathbf{x})\big]_{ij} \; = -\sum_{\mu=1}^M  \,  \, \frac{\partial^2 \ln P(\mu|\mathbf{x})}{\partial x_i \partial x_j} \; P(\mu|\mathbf{x})$. In the $K=1$-d case, it simply writes as: $I[\mu,\mathbf{r}] = I[\mu,x] - \frac{1}{2} \int \frac{F_{\text{cat}}(x)}{F_{\text{code}}(x)}\; P(x)\,dx$. The Fisher information $F_{\text{cat}}(x)$ characterizes the sensitivity of the category membership with respect to small variations of $x$. It is large at locations $x$ near a boundary between categories, and small if $x$ is well within a category. $F_{\text{code}}(x)$ is the `usual' Fisher information considered in neuroscience, related to the discriminability measured in psychophysics\citep{MacmillanCreelman91,seung1993simple}. It characterizes the sensitivity of the neural activity $\mathbf{r}$ with respect to small variations of $x$. The expression (\ref{eq:Delta_general}) allows for a simple and intuitive interpretation.\\
\textit{$\bullet$ Finding a proper discriminant space.} The first term, $I[\mu,\mathbf{x}]$, characterizes the correlation between the categories and the underlying projection space $X$. Maximizing this term means finding a discriminant space, an appropriate space from the point of view of the categorization task.\\
\textit{$\bullet$ Finding a proper metric.} The second term tells us what should be the metrics of the neural representation, how this space $X$ should be probed: the Fisher information $F_{\text{code}}$ should be large where the categorical Fisher information $F_{\text{cat}}$ is large in order to minimize the second term. Thus for a given space $X$, minimization of the second term in the mutual information leads to a neural code such that $F_{\text{code}}(x)$ is some increasing function of $F_{\text{cat}}(x)$ (see Appendix \ref{app:MinConstr}). Efficient coding in view of optimal classification is thus obtained by essentially matching the two metrics. Since $F_{\text{cat}}$ is larger near a class boundary, this should also be the case for $F_{\text{code}}(x)$. A larger $F_{\text{code}}(x)$ around a certain value of $x$ means that the neural representation is stretched at that location (the neural representation tiles the space $x$ more finely near than far from the class boundaries). Thus, category learning implies better cross-category than within-category discrimination, hence the so-called categorical perception. 

\section{Numerical experiments}
\label{sec:experiments}
\begin{figure}
	\centering
	\textbf{a.}\hspace{-0.3cm}
	\includegraphics[width=0.31\linewidth,valign=t]{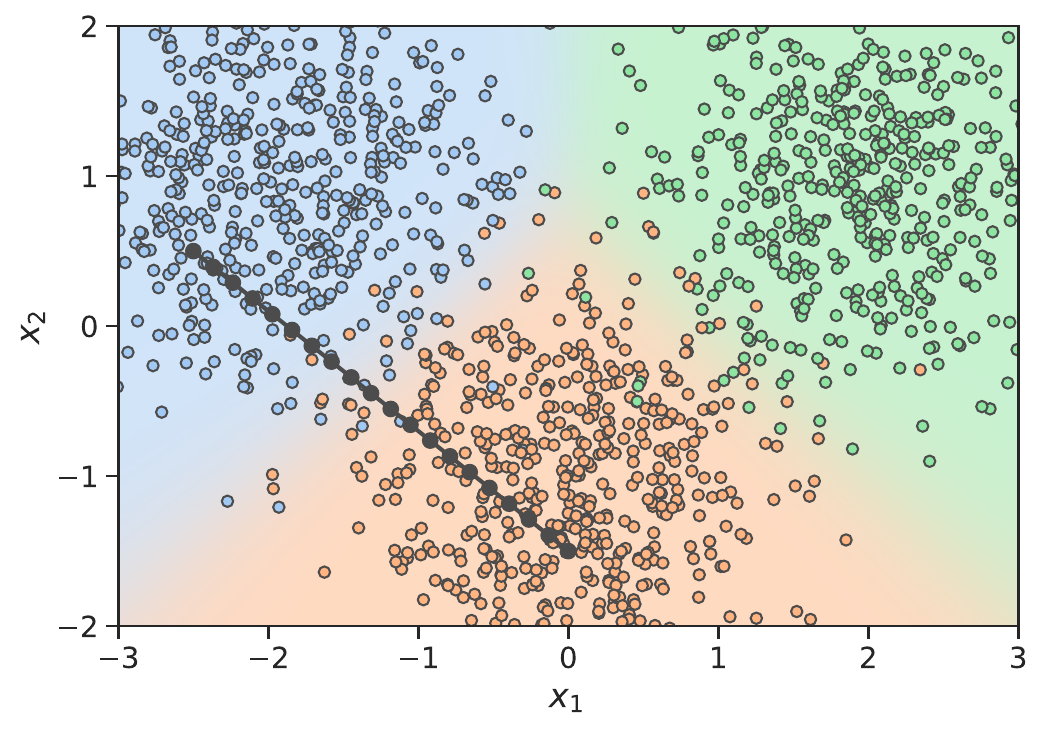}
	\hfill
	\textbf{b.}\hspace{-0.3cm}
	\includegraphics[width=0.31\linewidth,valign=t]{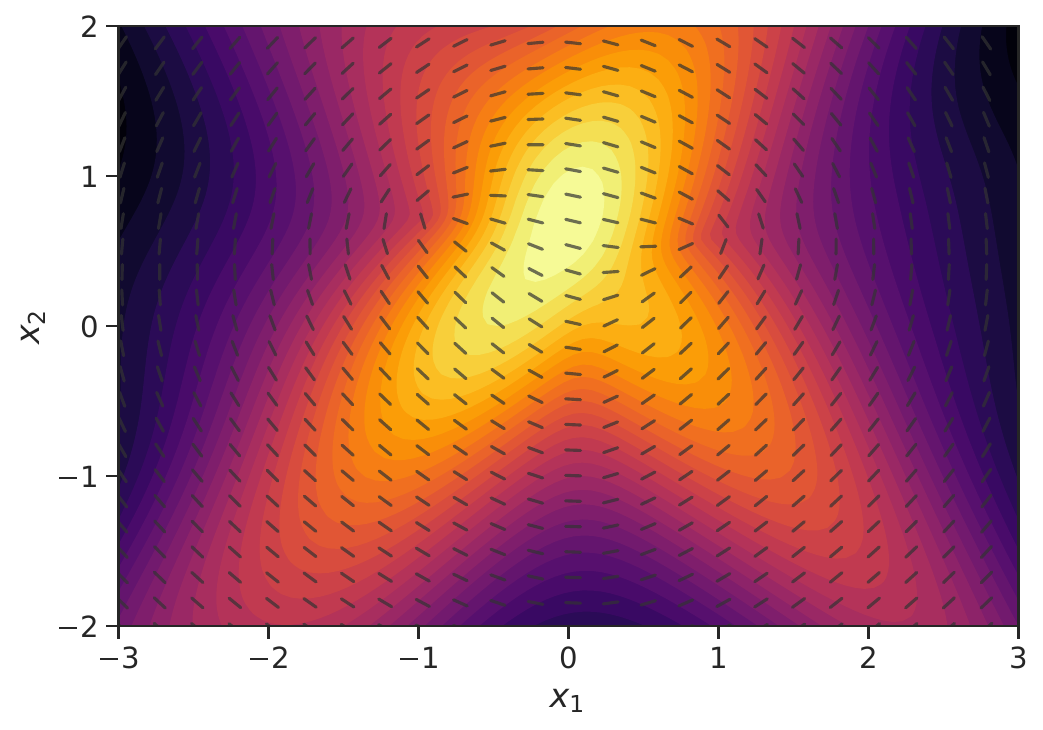}
	\hfill
	\textbf{c.}\hspace{-0.2cm}
	\includegraphics[width=0.34\linewidth,valign=t]{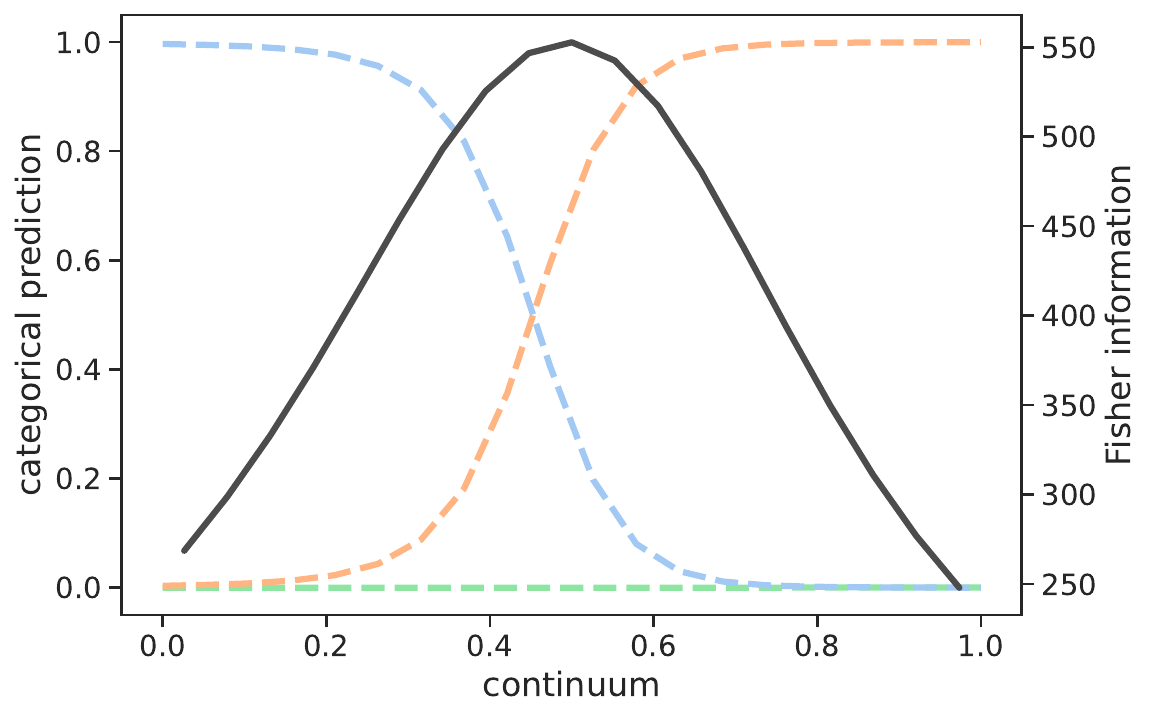}
	\caption{\textbf{Two-dimensional example with three Gaussian categories}. 
		(a) Colored dots: training set, random samples from each of the categories. Background color: mix between the colors that correspond to each of three categories, proportionally to the posterior probabilities $P(\mu|\mathbf{x})$ as estimated by the neural network.  Dark dots: a path interpolating between two samples from the blue and the red categories. 
		(b) Visualization of the Fisher information matrix at each point on the $(x_1, x_2)$ plane, after learning. The small line represents the direction at this point of the eigenvector of the Fisher information matrix associated with the largest eigenvalue. The magnitude of this largest eigenvalue is represented by the color, the lighter the greater.
		(c) The dotted colored lines indicate the posterior probabilities, as found by the network, each color representing its respective category. The solid line is the (scalar) Fisher information along the 1d path shown in (a).}
	\label{fig:gaussian2d}
\end{figure}

\paragraph{Two-dimensional example with Gaussian categories.}
We first consider a toy example involving a two-dimensional stimulus space with three overlapping Gaussian categories (see Fig.~\ref{fig:gaussian2d}a). Given the small dimension of $\mathbf{s}$, we work with $\mathbf{x} = \mathbf{s} \in\mathbb{R}^2$ (hence this $\mathbf{x}$ is given, not found by the network, $I[\mu,\mathbf{x}]$ and $F_{\text{cat}}(\mathbf{x})$ are here properties of the data). The neural network is a multilayer perceptron with one hidden layer of 32 cells. Each cell $i$ has a noisy neural activity given by $r_i(\mathbf{x}) = f_i(\mathbf{x}) + \sigma \sqrt{g_i(\mathbf{x})} z_i$, where $f_i$ is a sigmoidal activation function, $z_i$ is a unit normal random variable, and $\sigma=0.3$. Here we take $g_i(\mathbf{x}) = f_i(\mathbf{x})$. Note that this multiplicative noise can be seen as a form of dropout, which in the original work \citep{srivastava2014dropout} consists in multiplicative noise in the form of Bernoulli or Gaussian noise (where $g_i(\mathbf{x}) = f_i(\mathbf{x})^2$). The choice $g_i(\mathbf{x}) = f_i(\mathbf{x})$ yields a Poisson like noise, as commonly found in biological neural networks \citep{tolhurst1983statistical,softky1993highly}. We assume that the noise is not correlated between neurons given a stimulus $\mathbf{x}$, so that we can write $P(r|\mathbf{x}) = \prod P(r_i|\mathbf{x})$, which in turn helps writing the Fisher information as $F_{\text{code}}(\mathbf{x}) = \sum_i F_{\text{code}, i}(\mathbf{x})$, where $F_{\text{code}, i}(\mathbf{x})$ is the Fisher information of neuron $i$.\\
Figure~\ref{fig:gaussian2d}a shows that after learning the network has indeed learned to estimate the posterior probabilities $P(\mu|\mathbf{x})$, correctly partitioning the three categories into their respective regions. Figure~\ref{fig:gaussian2d}b presents a representation of the Fisher information $F_{\text{code}}(\mathbf{x})$ on the $\mathbf{x}$-plane after learning. Here, remember that the Fisher information is a $2\times2$ matrix. At each point, between classes, the eigenvector associated with the largest eigenvalue is orthogonal to the class boundary, and this eigenvalue is largest at the boundary between categories, illustrating the categorical perception phenomenon. Finally, we consider a 1d path in input space, depicted by the dark dots, interpolating between two items drawn from two different categories. We compute the scalar Fisher information of the neural code along this line. Figure~\ref{fig:gaussian2d}c shows the results, together with the categorical prediction outputted by the network. As expected, the neural Fisher information is the greatest at the boundary between categories.

\begin{figure}
	\centering
	\textbf{a.}\hspace{-0.3cm}
	\includegraphics[width=0.43\linewidth,valign=t]{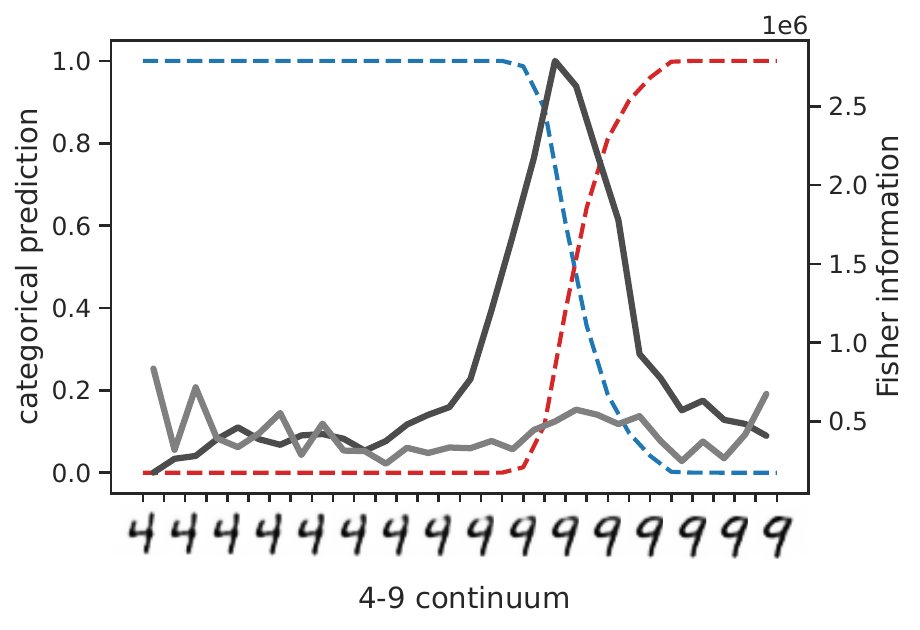}
	\hspace{0.2cm}
	\textbf{b.}\hspace{-0.3cm}
	\includegraphics[width=0.43\linewidth,valign=t]{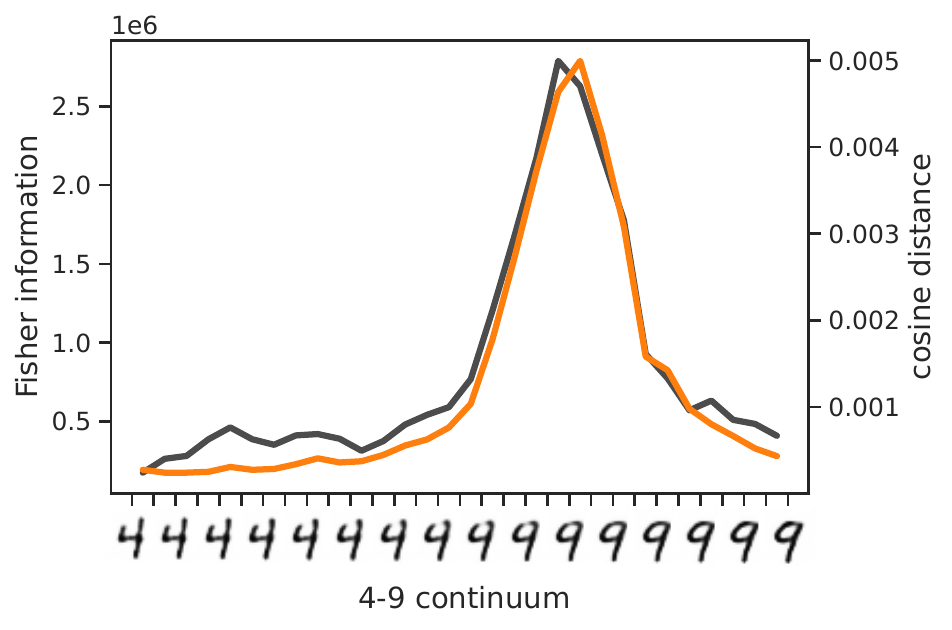}
	\caption{\textbf{Categorical perception along a 4 to 9 continuum}. (a) Neural Fisher information along the 4-9 continuum (average over 10 models), before (light gray) and after (dark gray) learning. Dotted colored lines:  posterior probabilities as found by the network, blue corresponding to category `4' and red to category `9'. (b) Comparison between Fisher information (dark gray) and cosine distance (orange, right y-axis) between neural activities evoked by contiguous items along the continuum.}
	\label{fig:mnist}
\end{figure}

\paragraph{Images of handwritten digits.}
Here we consider the MNIST dataset \citep{lecun1998gradient}, a large dataset of $28\times28$ handwritten digits (hence, the stimulus $\mathbf{s}$ lives in a 784 dimensional space). The neural network is a multilayer perceptron with two hidden layers, each made of 256 cells with ReLU activation. Poisson like neuronal noise affects the last hidden layer, just as in the previous example, with $\sigma=0.1$. A continuum between an item from the `4' category and an item from the `9' category (two categories that are among the most confusable ones) is created by interpolating between them in a latent space discovered by training an autoencoder to reconstruct digits from the MNIST training set \citep[as done in][]{LBG_JPN_CPDeepLearning_2022}. Each image along the continuum lies in the relevant manifold of digits. The labels in the x-axis of Fig.~\ref{fig:mnist}a pictures a few samples from the continuum, which is made of 31 images.
This continuum is considered as the 1d `$x$' in the previous discussions. One can then compute the categorical predictions outputted by the neural network together with the scalar Fisher information of the last hidden layer of neurons. Once again, Fig.~\ref{fig:mnist}a shows that learning induces categorical perception, with larger Fisher information at the boundary between the two categories. In a previous work \citep{LBG_JPN_CPDeepLearning_2022}, the cosine distance between the neural activities $\mathbf{r}(x)$ and $\mathbf{r}(x+\delta x)$ was used as a proxy for Fisher information $F_{\text{code}}(x)$, as it is much easier to compute. Fig.~\ref{fig:mnist}b shows that these two quantities indeed behave quite similarly.

\section{Discussion and challenges}
\label{sec:discussion}
We have shown that minimizing the mean Bayes cost in a categorization task notably implies maximizing the mutual information between category membership and neural activity. This optimization leads to (i) finding an appropriate representation space, and, (ii) building a representation with the appropriate metrics on this space, leading to an expansion of neural space near decision boundaries. The results presented here are based on previous works~\citep{LBG_JPN_2008,LBG_JPN_2012,LBG_JPN_CPDeepLearning_2022} and on a paper under preparation~\citep{LBG_JPN_Theory_2023}. \\
To conclude, we mention several challenging issues which should be addressed. (i) Our results are based on the use of the exact probability distributions of the data. They should be reconsidered in the context of learning with a finite set of examples. Note however that the numerical illustrations indicates that the main results hold in such a learning context. (ii) In the neuroscience context (but also in the machine learning context), one should study the effect of (possibly strong) noise at any stage of processing, also implying noise correlations in the subsequent layers -- in particular one issue is how to estimate the Fisher information quantity, $F_{code}$, as $P(\mathbf{r}|\mathbf{x})$ does not factorize in this case. (iii) It would be interesting to find ways of estimating the categorical Fisher information, $F_{cat}$. (iv) Finally an important issue is to understand the effect of noise correlations on the geometry of the neural space (in the spirit of \citep{franke2016structures}, but here for the case of category learning).

\clearpage
\bibliographystyle{plain}
\bibliography{refs.bib}

\clearpage
\renewcommand{\thesection}{Appendix \Alph{section}}
\renewcommand{\thesubsection}{\Alph{section}.\arabic{subsection}}
\setcounter{section}{0} 
\renewcommand{\theequation}{\Alph{section}.\arabic{equation}}
\setcounter{equation}{0} 
\renewcommand{\thefigure}{\Alph{section}.\arabic{figure}} 
\setcounter{figure}{0} 

\section{Supplementary Material}

\subsection{Link with the information bottleneck approach} 
\label{app:IB}
The information bottleneck (IB) approach \citep{tishby99information,tishby2000information} 
can be formulated as a rate distortion problem, the considered learning cost being  a distortion function that measures how well the category $\mu$ is predicted from the compressed neural representation $\mathbf{r}$ compared to its prediction from the stimulus $\mathbf{s}$. Tishby and collaborators developed this framework, theoretically and algorithmically, in the context of deep learning \citep{tishby-zaslavsky2015,schwartz-ziv2017opening}.

The qualitative idea of the IB approach is that the neural activity should convey as little information as possible about the stimulus provided the information about the category is preserved. Thus, with our notation, the goal is to minimize $I[\mathbf{s},\mathbf{r}] - \beta I[\mu,\mathbf{r}]$ where $\beta$ is a Lagrange multiplier. Analyzing this optimization principle, Tishby {\it et al} \citep{tishby2000information} show that the Kullback-Leibler divergence $D_{KL}(P_{\mu|\mathbf{s}}||P_{\mu|\mathbf{r}})$ emerges as the relevant effective distorsion measure. This divergence corresponds to our cost function once the decoding stage is optimized, that is $g_{\mu}(\mathbf{r}) = P(\mu|\mathbf{r})$. Then one sees that the approach followed here is somewhat dual to the IB one. One starts from the K-L divergence, and the infomax criterion emerges from the cost function. There is however two important differences. First, the full cost function that we consider includes the decoding part, and second, the correspondence is with the IB cost in the $\beta \rightarrow \infty$ limit.\\
An alternative way to see this correspondence is to consider, from a distortion measure viewpoint,  the IB cost associated to the Bayes cost:
\begin{equation*}
\mathcal{\overline{C}}_{IB}(\beta) = I[\mathbf{s},\mathbf{r}] + \beta \, \mathcal{\overline{C}}
\label{eq:cost_ib}
\end{equation*}
Making use of the decomposition of $\mathcal{\overline{C}}$ in coding and decoding parts (Eq.~\ref{eq:cost_main_expression}), we can write
\begin{equation*}
\mathcal{\overline{C}}_{IB}(\beta)  
= \mathcal{\overline{C}}_{IB,coding}(\beta) + \beta \, \mathcal{\overline{C}}_{decoding}
\label{eq:cost_main_expression_bis}
\end{equation*}
where
\begin{equation}
\mathcal{\overline{C}}_{IB,coding}(\beta) \,=\, I[\mathbf{s},\mathbf{r}] + \beta\, \left( I[\mu,\mathbf{s}] - I[\mu,\mathbf{r}] \right)
\label{eq:Ccod_IB} 
\end{equation}
Since $I[\mu,\mathbf{s}]$ is a constant, $\mathcal{\overline{C}}_{IB,coding}(\beta)$ 
is the usual information bottleneck cost function.

\subsection{Minimization under constraints}
\label{app:MinConstr}
We comment here on the minimization of the part of the coding cost which depends on the Fisher information quantities, $F_{\text{code}}$ and $F_{\text{cat}}$, for a given projection space $X$ -- hence a given categorical information $F_{\text{cat}}$.
As explained in the main text, minimization of this term requires that $F_{\text{code}}$ essentially follows the categorical Fisher information $F_{\text{cat}}$. The precise result will depend on the constraints on the neural system. 
The constraints may be on the neurons parameters, as in \cite{KB_JPN_2020}, or directly on the Fisher information considered as a function, as in \cite{LBG_JPN_2008}.  In such case one minimizes the right hand side of equation (\ref{eq:Delta_general}) under the chosen constraint $\Psi{}$ (for simplicity we consider the 1d case),
\begin{eqnarray}
\mathcal{E} &=&\frac{1}{2} \int_X  \frac{F_{\text{cat}}(x)}{F_{\text{code}}(x)}\; P(x)\,dx %\nonumber \\
% &+& 
\;+\; 
\lambda \left(\int_X \Psi(F_{\text{code}}(x))\;P(x)\,dx \;-\;c \,\right)
\label{eq:optFishPsi}
\end{eqnarray}
For instance, if $\Psi(F)=F^{\alpha}$, one gets $F_{\text{code}}(x) \propto [F_{\text{cat}}(x)]^{\frac{1}{1+\alpha}}$, which is meaningful for $\alpha>1$. The limit $\alpha \rightarrow 0$  corresponds to considering an information theoretic constraint, as we show now.

As presented Section \ref{app:IB}, adopting the viewpoint of the information bottleneck approach~\citep{tishby99information}, we may minimize the mutual information $I[x,\mathbf{r}]$
under the constraint that the information conveyed by the neural code about the categories is large enough:
\begin{eqnarray}
\mathcal{E} &=& I[x,\mathbf{r}] - \beta I[\mu,\mathbf{r}] \nonumber
\label{eq:bottleneck_1}
\end{eqnarray}
In the same asymptotic limit as the one considered here,  $I[x,\mathbf{r}]$ behaves as $\frac{1}{2} \int \ln F_{\text{code}}(x)\;P(x)\,dx$ (again here for $K=1$) \citep{Brunel_Nadal_1998} . Combining this result and the ones in  \citep{LBG_JPN_2008}, we can thus write 
\begin{eqnarray}
\mathcal{E} &=& \frac{1}{2} \int \ln F_{\text{code}}(x)\;P(x)\,dx \nonumber \nonumber \\
&-& \beta \left(I[\mu,x] - \frac{1}{2} \int_X  \frac{F_{\text{cat}}(x)}{F_{\text{code}}(x)}\; P(x)\,dx\right) \nonumber
\label{eq:bottleneck_2}
\end{eqnarray}
Up to the (here constant) term $I[\mu,x]$, this is equivalent to the cost (\ref{eq:optFishPsi}), in the case $\Psi(.)=\ln(.)$, taking the dual approach -- that is exchanging the roles of the cost and the constraint, $\beta=1/\lambda$. The optimal function is here $F_{\text{code}}(x) \propto F_{\text{cat}}(x)$.

\end{document}